\documentclass{article}

\usepackage{PRIMEarxiv}

\usepackage[utf8]{inputenc} 
\usepackage[T1]{fontenc}    
\usepackage{hyperref}       
\usepackage{url}
\usepackage{amsmath}
\usepackage{booktabs}       
\usepackage{amsfonts}       
\usepackage{nicefrac}       
\usepackage{microtype}      
\usepackage{lipsum}
\usepackage{fancyhdr}       
\usepackage{graphicx}       
\graphicspath{{images/}}     

\usepackage{subcaption} 

\usepackage{xcolor}%

\title{Kolmogorov-Arnold Network Autoencoders in medicine.
}

\author{
  Ugo Lomoio \\
  Department of surgical and medical sciences \\
  Magna Graecia University of Catanzaro \\
  Catanzaro, Italy, 88100 \\
\texttt{ugo.lomoio@unicz.it} \\
  \And 
  Pierangelo Veltri \\
  DIMES \\
  University of Calabria \\
  Rende, Italy, 87036 \\
\texttt{pierangelo.veltri@dimes.unical.it } \\
  \And
  Pietro Hiram Guzzi \\
  Department of surgical and medical sciences \\
  Magna Graecia University of Catanzaro \\
  Catanzaro, Italy, 88100 \\
  \texttt{hguzzi@unicz.it} \\
}

\begin{document}
\maketitle

\begin{abstract}

Deep learning neural networks architectures such Multi Layer Perceptrons (MLP) and Convolutional blocks still play a crucial role in nowadays research advancements.  From a topological point of view, these architecture may be represented as graphs in which we learn the functions related to the nodes while fixed edges convey the information from the input to the output. A recent work introduced a new architecture called Kolmogorov Arnold Networks (KAN) that reports how putting learnable activation functions on the edges of the neural network leads to better performances in multiple scenarios. Multiple studies are focusing on optimizing the KAN architecture by adding important features such as dropout regularization, Autoencoders (AE), model benchmarking and last, but not least, the KAN Convolutional Network (KCN) that introduced matrix convolution with KANs learning. This study aims to benchmark multiple versions of vanilla AEs (such as Linear, Convolutional and Variational) against their Kolmogorov-Arnold counterparts that have same or less number of parameters. Using cardiological 
signals as model input, a total of five different classic AE tasks were studied: reconstruction, generation, denoising, inpainting and anomaly detection. 
The proposed experiments uses a medical dataset \textit{AbnormalHeartbeat} that contains audio signals obtained from the  stethoscope. 
\end{abstract}

\keywords{autoencoder \and KAN \and medicine \and signal analysis}

\section{Introduction}
\label{sec:intro}

Deep learning architectures based on Multi-Layer Perceptrons (MLPs) have served as foundational tools in driving major breakthroughs across numerous scientific fields, including bioinformatics and medical informatics \cite{lecun2015deep,guzzi2020biological}.

In the standard formulation, an MLP can be seen as a computation graph where the nodes correspond to learnable parameters (synaptic weights and biases), while the edges represent fixed, identical activation functions (such as ReLU or Tanh) applied across each layer. This structure ensures that the network maps input data through layers of transformations using the same nonlinearity at every connection, with the learning process focused solely on optimizing the node parameters.

\begin{figure}[ht]
    \centering
    \includegraphics[width=0.8\linewidth]{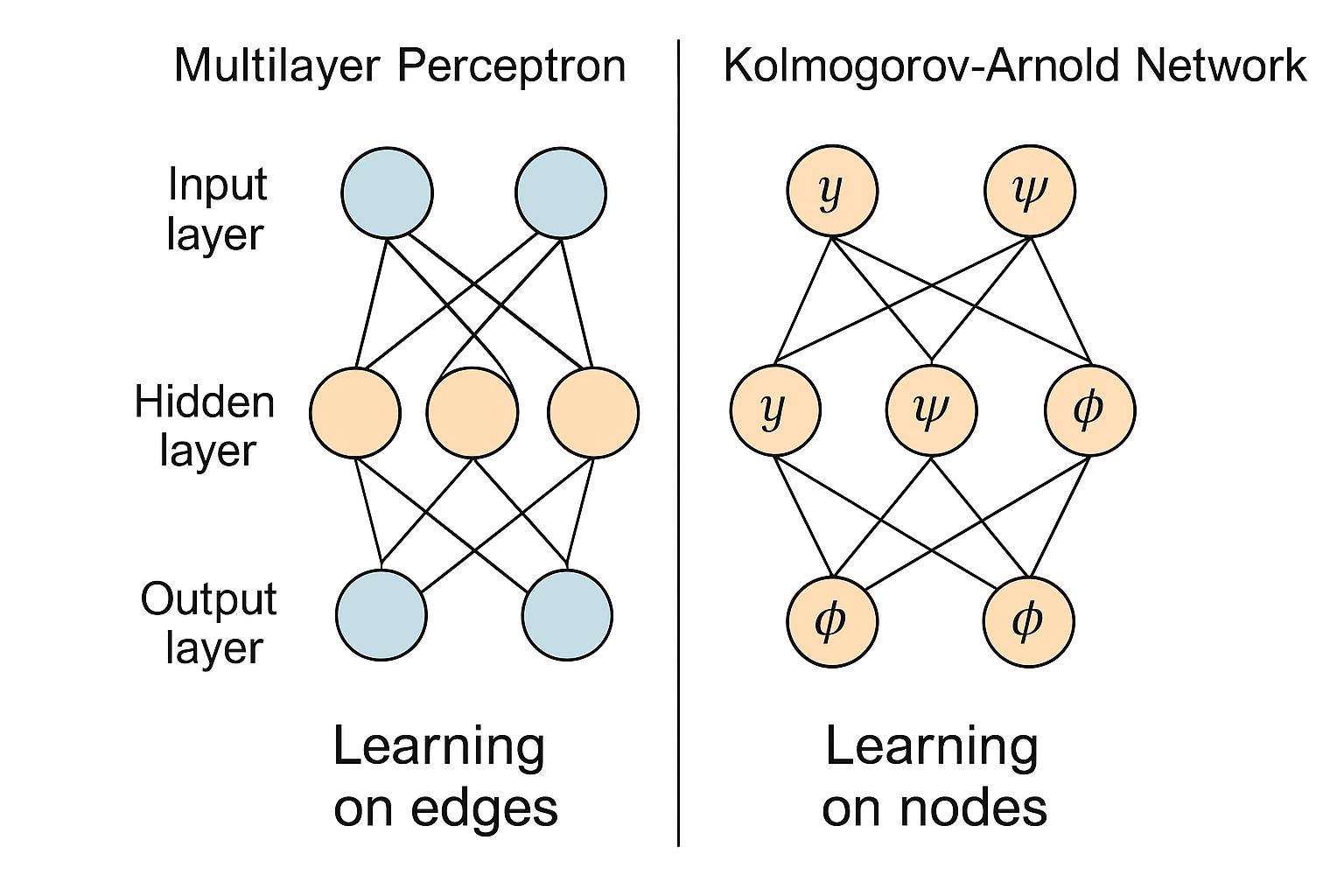}
     \caption{\textbf{Topological comparison between a Multilayer Perceptron (MLP) and a Kolmogorov--Arnold Network (KAN).}
    The MLP (left) consists of fully connected layers where learning occurs via scalar weights on edges between nodes, enabling the network to approximate complex functions by tuning edge connections. In contrast, the KAN (right) replaces scalar edge weights with learnable univariate functions at each node, inspired by the Kolmogorov--Arnold representation theorem. Here, learning is concentrated within nodes themselves, each applying a nonlinear transformation (denoted as $(\psi)$, $\phi$, and $y$) to their inputs. This architectural distinction shifts the burden of expressivity from edge weights to functional transformations at nodes, potentially enhancing interpretability and reducing parameter count.
    }
    \label{fig:kanvsmlp}
\end{figure}

Recent developments have led to a reconsideration of this model. One particularly novel approach is embodied by Kolmogorov–Arnold Networks (KANs), which offer a fundamental change of paradigm. Rather than assigning fixed activation functions to each node, KANs replace them with learnable, adaptive functions placed directly on the edges of the network graph \cite{kan2024arxiv,lomoio2025design}. This edge-wise parametrization grants the network greater expressive power and flexibility, enabling it to model intricate and domain-specific data patterns better. As illustrated in Figure \ref{fig:kanvsmlp}, this represents a significant departure from the traditional architecture of MLPs and CNNs, which could open the door to more accurate and interpretable models in fields where capture of complex relationships is paramount.

This innovative approach draws its theoretical foundation from the Kolmogorov–Arnold representation theorem \cite{kolmogorov1957representation}, which establishes that any continuous multivariate function can be exactly represented as a finite superposition of continuous univariate functions and addition operations. Leveraging this mathematical result, Kolmogorov–Arnold Networks (KANs) adopt an edge-centric perspective on neural computation by assigning learnable, univariate activation functions directly to the network edges. This formulation enables KANs to approximate complex multivariate mappings through compositions of simpler, adaptively learned univariate transformations, resulting in richer expressive capacity and more flexible adaptation to diverse tasks compared to traditional architectures that use fixed node-based nonlinearities \cite{kan2024arxiv}.

The initial formulation of KANs has since been extended to represent many classical architecture of neural networks such as  Autoencoders (AEs) \cite{kanconv2024,hiram2022disease}. These enhancements have opened the door to applying KANs to unsupervised learning settings and spatially structured data, making them a promising candidate for applications in biomedical signal analysis.

Autoencoders are a class of neural networks used for \textbf{unsupervised representation learning} \cite{lomoio2025convolutional}, particularly for tasks such as \textit{dimensionality reduction}, \textit{denoising}, and \textit{generative modeling}. 
A \textbf{vanilla autoencoder} is the simplest form of this architecture and consists of two main components:

\begin{itemize}
    \item \textbf{Encoder}: A function \( f_{\text{enc}}: \mathbb{R}^n \rightarrow \mathbb{R}^k \) that compresses input data \( x \in \mathbb{R}^n \) into a latent representation \( z \in \mathbb{R}^k \) (with \( k < n \)).
    \item \textbf{Decoder}: A function \( f_{\text{dec}}: \mathbb{R}^k \rightarrow \mathbb{R}^n \) that reconstructs the original input from the latent representation, producing \( \hat{x} \).
\end{itemize}

The model is trained to minimise a reconstruction loss, typically the mean squared error (MSE):

\[
\mathcal{L}(x, \hat{x}) = \|x - \hat{x}\|^2
\]

In traditional implementations, both encoder and decoder are modelled using \textbf{multi-layer perceptrons (MLPs)} composed of fully connected layers and nonlinear activation functions such as ReLU or Tanh.
Although KANs have shown promise in supervised learning tasks, their ability to learn unsupervised representation generally carried out by AE remains underexplored. 

Here we systematically compare KAN  and MLP-based autoencoders, evaluating their performance across multiple datasets and regimes. We analyse reconstruction fidelity, latent space structure, and interpretability, intending to clarify the practical and theoretical implications of node-centric learning in functional neural architectures.

As an applicative scenario, we use the  \texttt{AbnormalHeartbeat} dataset, a corpus of stethoscope-derived recordings. We consider five canonical AE tasks: reconstruction, generation, denoising, inpainting, and anomaly detection. For each task, we implement multiple variants of autoencoders with similar or fewer parameters in the Kolmogorov versions to ensure fair benchmarking.

Our empirical findings show that MLP-based autoencoders—whether classical or KAN-based—struggle with medical signal representation. In contrast, convolutional autoencoders perform more robustly, with KAN-convolutional autoencoders achieving superior performance across all tasks. These results highlight the representational benefits of edge-learnable activations in structured data domains.

Despite their advantages, KAN-based models currently incur a significant computational cost: the forward pass in a Kolmogorov layer is up to 10x slower than that of a conventional MLP layer, and notably slower than CNN layers. This tradeoff between accuracy and efficiency presents a compelling direction for future optimization and hardware-aware implementation.

\section{Background}
\label{sec:back}
Biomedical signals—such as electrocardiograms (ECG), electroencephalograms (EEG), and electromyograms (EMG)—capture complex physiological dynamics but are often challenged by high dimensionality, non-stationary noise, and inter-subject variability \cite{zitnik2024current,hiram2022disease,mercatelli2021exploiting}. Autoencoders (AEs), a class of unsupervised neural networks, provide a powerful framework for extracting meaningful representations from such data by compressing signals into compact latent spaces and reconstructing them with minimal information loss \cite{hinton2006reducing,lomoio2025convolutional}. These latent representations can be leveraged for denoising, anomaly detection, and feature extraction in downstream analytical tasks.

An AE architecture generally comprises two components: an encoder that transforms the input signal into a low-dimensional latent code, and a decoder that reconstructs the original input from this code. Extensions of the basic model—such as denoising autoencoders (DAEs) and variational autoencoders (VAEs)—further enhance robustness and interpretability. DAEs are specifically designed to reconstruct clean signals from corrupted inputs, thereby improving noise resilience. VAEs introduce a probabilistic framework that enables generative modeling with structured latent spaces, supporting uncertainty quantification and data synthesis \cite{kingma2013auto}.

Among biomedical domains, ECG signal processing has witnessed extensive adoption of AE-based techniques. ECG recordings are susceptible to artifacts arising from muscular activity, electrode displacement, and electromagnetic interference. Conventional preprocessing methods—including digital filtering and wavelet decomposition—often rely on handcrafted heuristics, which may lack adaptability to diverse noise profiles. In contrast, DAEs offer a data-driven approach to suppress noise while preserving diagnostically relevant signal characteristics. For example, convolutional autoencoders have been successfully trained to reconstruct clean ECG signals and extract features instrumental for myocardial infarction detection \cite{zhai2017automated}. Acharya et al. demonstrated that deep convolutional architectures can automatically identify congestive heart failure from single-lead ECG traces, achieving high diagnostic accuracy without manual feature engineering \cite{acharya2017deep}.

In practical clinical scenarios, AEs have also been integrated into real-time arrhythmia detection pipelines \cite{yildirim2018arrhythmia}. By encoding ECG waveforms into low-dimensional latent vectors, these models facilitate unsupervised clustering and outlier detection, enabling the identification of abnormal beats such as premature ventricular contractions. The resulting latent embeddings support patient-specific modeling, thus paving the way toward personalized diagnostics.

Variational autoencoders further expand the utility of AEs by modeling latent variables probabilistically. Instead of deterministic encoding, VAEs learn a distribution—usually Gaussian—over the latent variables, which permits not only signal reconstruction but also sample generation, interpolation, and uncertainty estimation \cite{kingma2013auto}. In ECG applications, VAEs have been applied to data augmentation and quality prediction. Conditional VAEs, for example, can generate physiologically realistic ECG segments conditioned on specific waveform classes or noise profiles, addressing challenges of limited labeled data \cite{li2021conditional}. Moreover, VAEs trained exclusively on normal ECG signals can effectively detect anomalies by identifying increased reconstruction errors or deviations from the learned latent distribution \cite{an2015variational}. These features are particularly valuable in continuous monitoring environments where early detection of rare cardiac events is critical.

Notably, the structured latent space learned by VAEs enhances interpretability, allowing clinicians and researchers to correlate latent dimensions with physiological factors such as heart rate variability or morphological alterations in the QRS complex. This interpretability is essential for bridging the gap between black-box deep learning models and clinical decision-making, thereby fostering transparency and trust in AI-assisted diagnostics.

In summary, autoencoders and their variational extensions constitute a flexible, data-efficient, and interpretable toolkit for modeling biomedical signals. Their demonstrated success in ECG analysis—ranging from denoising to anomaly detection—underscores the growing potential of unsupervised deep learning methods to address the challenges posed by increasingly large, heterogeneous, and dynamic clinical datasets.

\subsection{Kolmogorov Arnold Networks}

Kolmogorov–Arnold Networks (KANs) are a novel class of function-approximating neural networks inspired by the  Kolmogorov–Arnold representation theorem . Unlike traditional multilayer perceptrons (MLPs) which rely on compositions of affine transformations and fixed activation functions, KANs replace the linear weights of standard networks with learnable  univariate functions , thereby aligning more directly with the theoretical decomposition of multivariate functions.

KANs are not merely architectural alternatives but are grounded in the fundamental theory of function representation, providing a powerful tool for interpretable and sparse approximations of smooth or structured functions.

\paragraph{Kolmogorov–Arnold Theorem}

\begin{figure}
    \centering
    \includegraphics[width=0.8\linewidth]{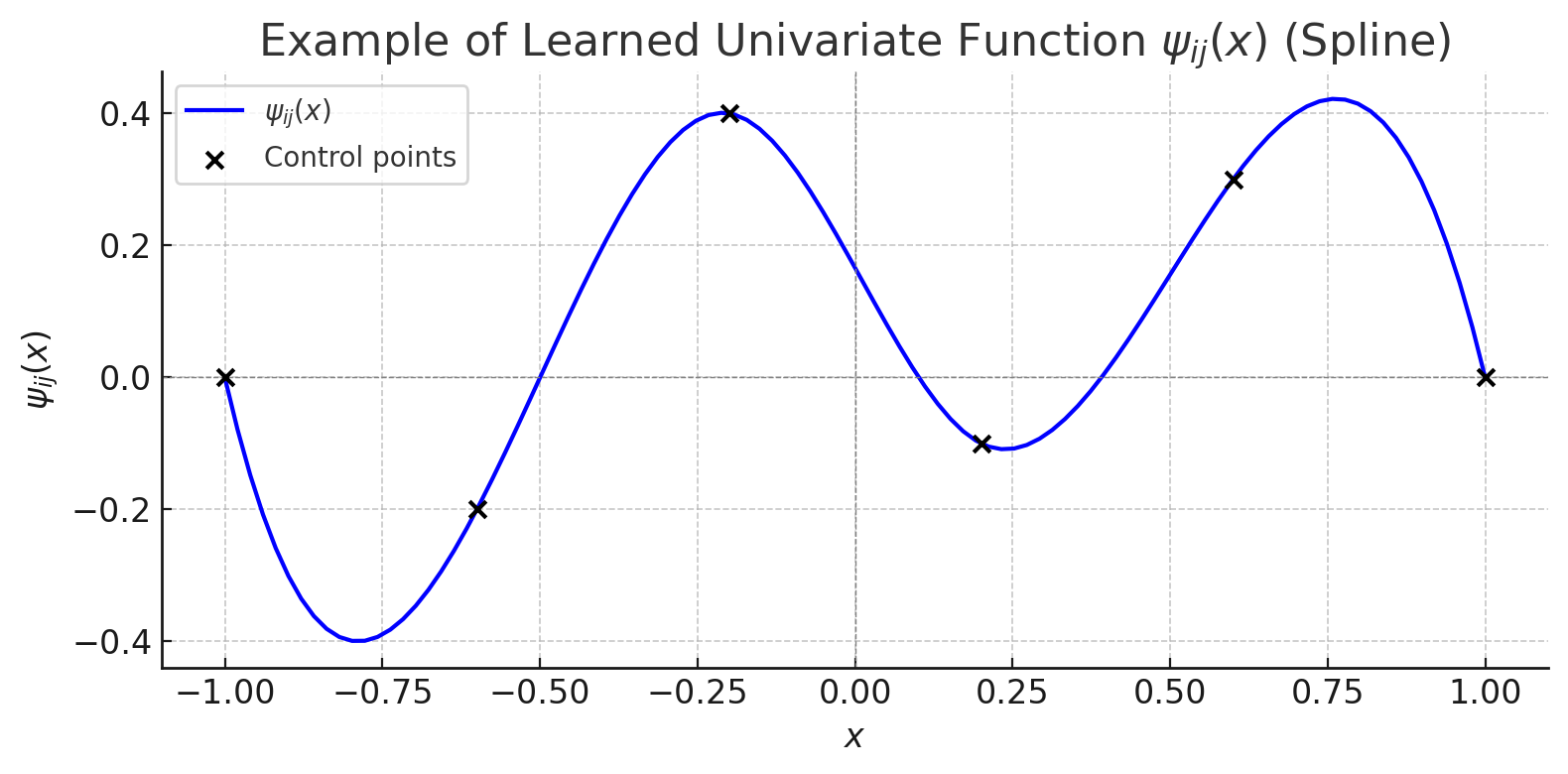}
    \caption{Example of a learned univariate function $\psi_{ij}(x)$ in a Kolmogorov--Arnold Network, modeled as a smooth spline.}
\label{fig:psi_spline}
\end{figure}

The Kolmogorov–Arnold theorem asserts that any continuous function \( f : [0,1]^n \rightarrow \mathbb{R} \) can be represented as a finite sum of compositions of continuous univariate functions as depicted in Figure \ref{fig:psi_spline}. Specifically:

\textbf{Theorem (Kolmogorov, 1957)} \cite{kolmogorov1957representation, arnold1957functions}:
For every continuous function \( f : [0,1]^n \rightarrow \mathbb{R} \), there exist continuous functions \( \phi_q : \mathbb{R} \rightarrow \mathbb{R} \) and \( \psi_{qj} : \mathbb{R} \rightarrow \mathbb{R} \), where \( 1 \leq q \leq 2n+1 \) and \( 1 \leq j \leq n \), such that
\begin{equation}
f(x_1, \ldots, x_n) = \sum_{q=1}^{2n+1} \phi_q \left( \sum_{j=1}^n \psi_{qj}(x_j) \right).
\label{eq:kolmogorov}
\end{equation}

This decomposition shows that multivariate functions can be expressed without using multivariate nonlinearities, by relying solely on univariate functions and additions. This result laid the mathematical foundation for the construction of Kolmogorov–Arnold Networks.

\paragraph{Definition of Kolmogorov–Arnold Networks}

\textbf{Definition:} A Kolmogorov–Arnold Network (KAN) is a neural network model where each connection between nodes in consecutive layers is not associated with a scalar weight but rather a learnable univariate function. The network approximates a target function \( f : \mathbb{R}^n \to \mathbb{R}^m \) as a sum of composed univariate mappings, inspired by Equation~\ref{eq:kolmogorov}.

\begin{figure}[ht]
\centering
\includegraphics[width=0.8\textwidth]{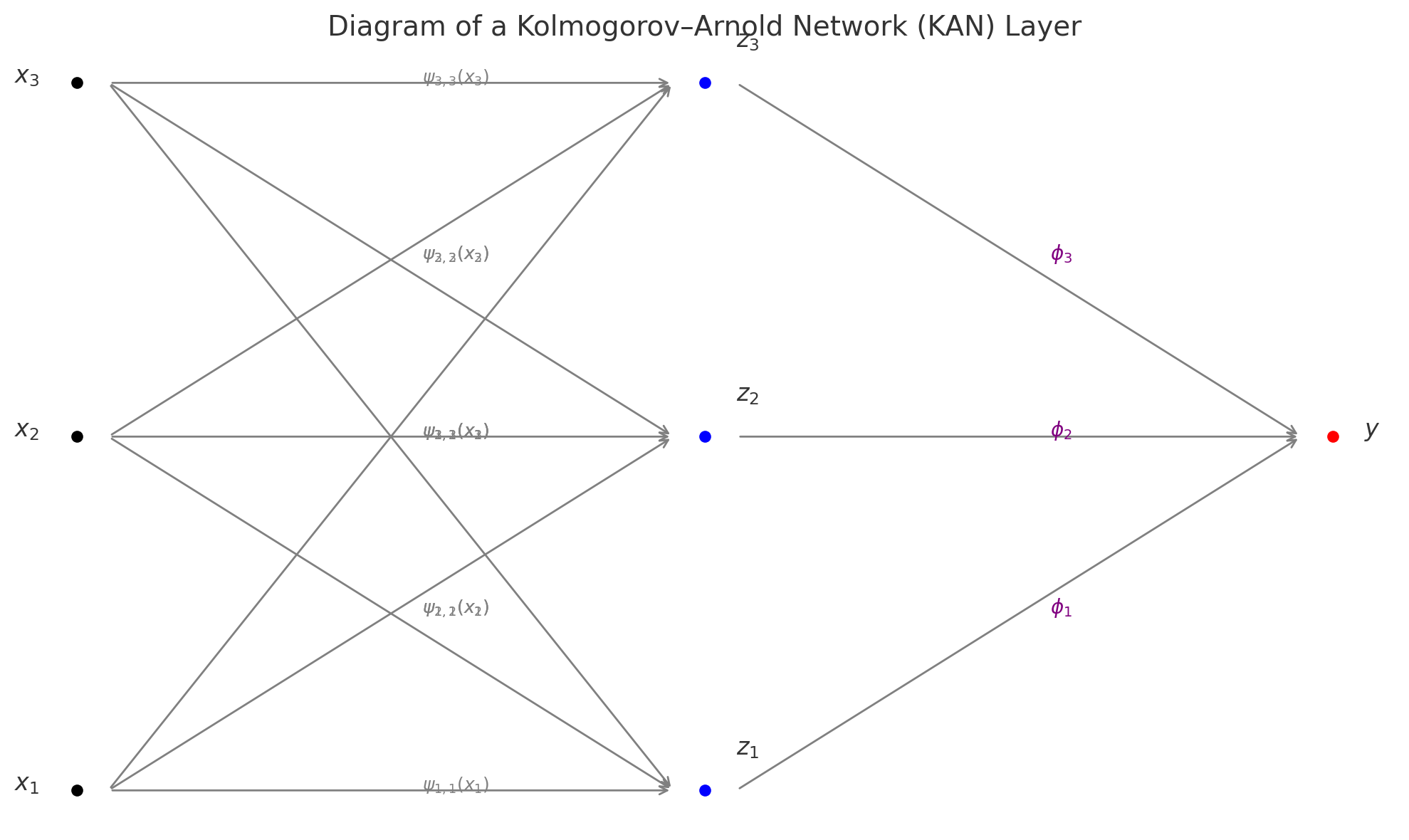} 
\caption{
Diagram illustrating a single layer of a Kolmogorov--Arnold Network (KAN). 
Each edge from an input node to a hidden node applies a learned univariate function 
$\psi_{ij}(x_j)$. Each hidden node sums its incoming transformed values and applies 
a univariate function $\phi_i$. The output node aggregates the results from the hidden nodes.
}
\label{fig:kan-layer}
\end{figure}

\textbf{Formally:}, as depicted in Figure \ref{fig:kan-layer},  In a KAN layer, the transformation is defined as:
\begin{equation}
z_i = \sum_{j=1}^{n} \psi_{ij}(x_j), \quad y_i = \phi_i(z_i),
\label{eq:kan_layer}
\end{equation}
where:
\begin{itemize}
    \item \( x_j \) is the input to node \( j \),
    \item \( \psi_{ij} : \mathbb{R} \rightarrow \mathbb{R} \) is a learnable univariate function associated with the edge from node \( j \) to node \( i \),
    \item \( \phi_i : \mathbb{R} \rightarrow \mathbb{R} \) is an activation function for node \( i \),
    \item \( y_i \) is the output of node \( i \) in the current layer.
\end{itemize}

Multiple such layers can be stacked to approximate complex functions. Unlike MLPs, the structure of KANs allows each path through the network to preserve univariate compositionality.

In practice, each \( \psi_{ij} \) and \( \phi_i \) can be parameterized using:
\begin{itemize}
    \item  Piecewise polynomials  (e.g., cubic splines),
    \item  Fourier or Chebyshev expansions ,
    \item  Small neural subnets  (MLPs) restricted to single input variables.
\end{itemize}

The parameters of these basis expansions are learned via backpropagation. Regularization terms are often applied to encourage smoothness and to constrain the function range or derivatives.

\section{Methods}
\label{sec:methods}
\subsection{Datasets}
\label{sec:datasets}

In this work the UCR AbnormalHeartbeat dataset datatset has been used \cite{ucrarchive}. The dataset is composed of a total of 1,189 time series, partitioned into a training set and a testing set. The training set contains 100 instances (50 from each class), ensuring balanced class representation for training. The testing set consists of 1,089 instances, with an inherent class imbalance reflecting more realistic clinical distributions. 

\textbf{Dataset statistics:}
\begin{itemize}
    \item \textbf{Number of classes:} 2 (\textit{normal}, \textit{abnormal}).
    \item \textbf{Number of training instances:} 100 (50 per class).
    \item \textbf{Number of testing instances:} 1,089.
    \item \textbf{Length of each time series:} 187 samples.
\end{itemize}

The AbnormalHeartbeat dataset is commonly used as a benchmark for time series classification algorithms in the biomedical domain. Its small training set and imbalanced test distribution pose challenges that test the generalization ability of machine learning models, making it a suitable testbed for both traditional and deep learning-based approaches. The clinical relevance of distinguishing abnormal from normal heartbeats further underlines the importance of effective feature learning and robust classification strategies.







\begin{figure}[ht]
    \centering
    \includegraphics[width=0.5\linewidth]{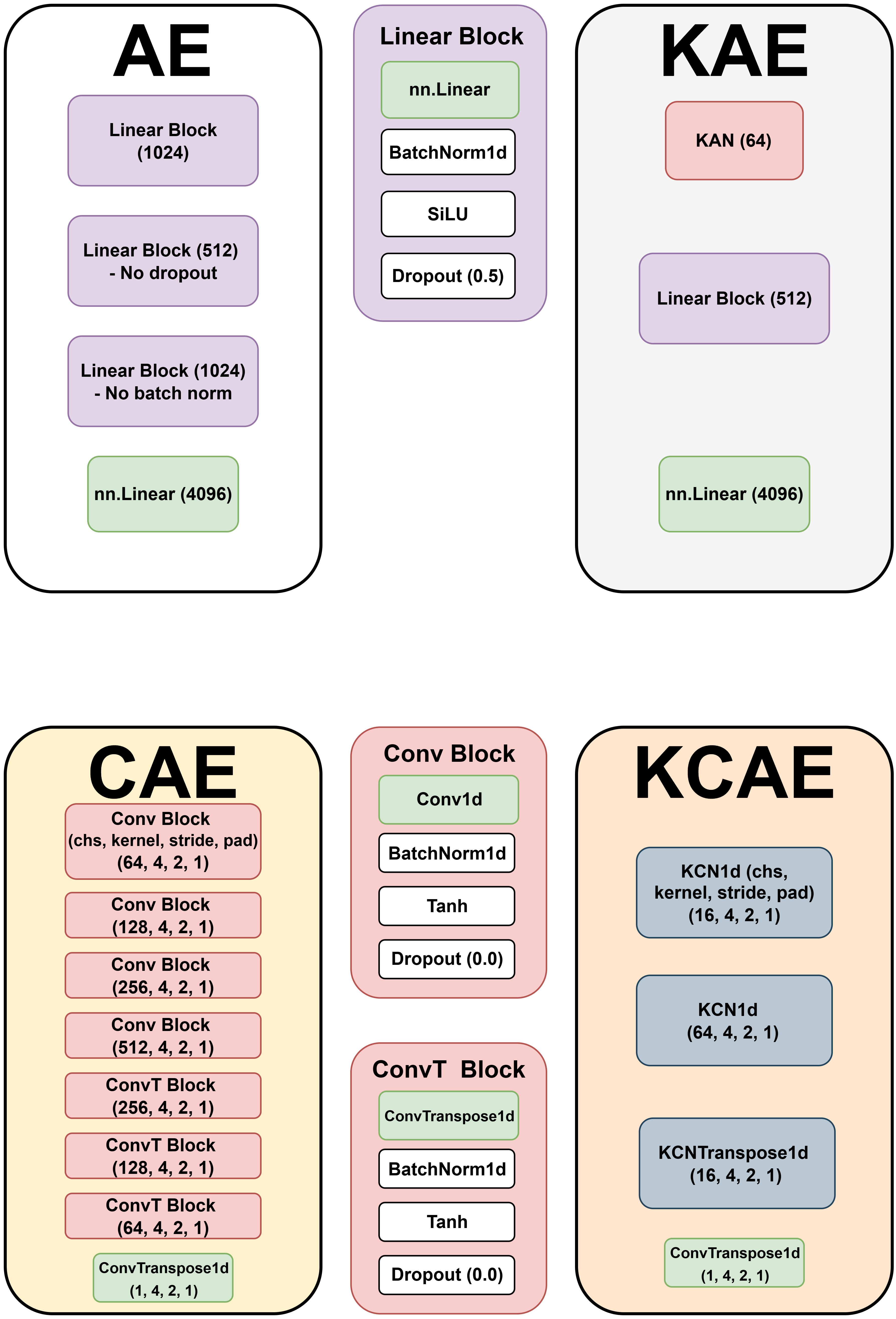}
    \caption{Architectures}
    \label{fig:architectures}
\end{figure}

In this work, we investigate four different autoencoder (AE) architectures, each designed to process input data through distinct architectural paradigms. Figure~\ref{fig:architectures} summarizes the main components of each architecture.

\textbf{AutoEncoder (AE):} The AE model employs fully connected (dense) layers organized into sequential \textit{Linear Blocks}, each comprising a linear transformation followed by batch normalization, a SiLU activation function, and dropout regularization. Specific layers omit dropout or batch normalization to assess their individual contributions to the network's performance. The AE concludes with a linear output layer.

\textbf{Kernel-based AutoEncoder (KAE):} The KAE modifies the standard AE by introducing \textit{Kernel Activation Network (KAN)} layers, which allow for more expressive nonlinear transformations. This architecture replaces the initial linear block with a KAN block while maintaining subsequent linear processing layers.

\textbf{Convolutional AutoEncoder (CAE):} The CAE replaces fully connected layers with one-dimensional convolutional layers, making it particularly suitable for sequential data such as time series. It uses \textit{Conv Blocks} composed of Conv1d layers, batch normalization, Tanh activation, and dropout. The decoder part uses \textit{ConvTranspose1d} layers to upsample the latent representations back to the input dimension.

\textbf{Kernel-based Convolutional AutoEncoder (KCAE):} The KCAE extends the CAE by introducing \textit{Kernel-based Convolutional Network (KCN1d)} layers, combining convolutional operations with kernel-based nonlinear activations. This architecture captures both local patterns and complex nonlinear relationships within the data. Similar to CAE, it employs convolutional transpose layers in the decoder stage.

Overall, these architectures enable a comparative analysis of standard versus kernel-based approaches in both fully connected and convolutional frameworks, providing insights into their reconstruction capabilities and feature extraction performance.

\section{Results and Discussion}
\label{sec:results}

\subsection{Experimental Results}

In this study, we evaluate and compare the reconstruction performance of two autoencoder architectures: a standard Convolutional Autoencoder (CAE) and a Kernel-based Convolutional Autoencoder (KCAE), applied to the AbnormalHeartbeat dataset. The key objective is to assess how effectively each model encodes class-specific features into a compressed latent representation. To this end, we analyze the learned latent spaces using Uniform Manifold Approximation and Projection (UMAP), which provides a two-dimensional visualization of the latent embeddings corresponding to normal and abnormal heartbeat classes during the reconstruction task.

\begin{figure}[ht]
\vskip\baselineskip
\begin{subfigure}{0.49\linewidth}
    \centering
    \includegraphics[width=\linewidth, keepaspectratio]{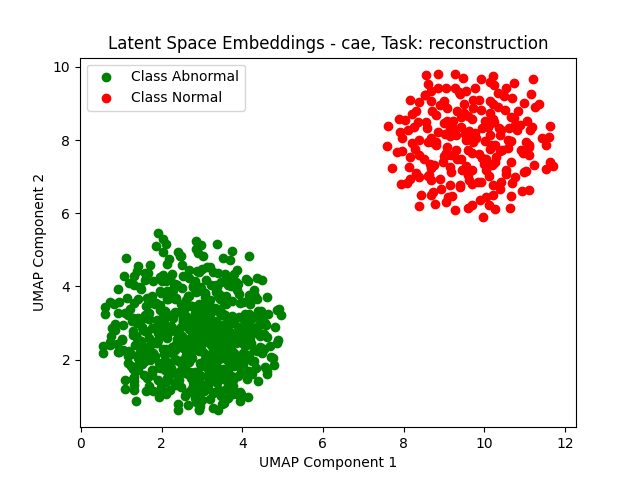}
    \caption{}
    \label{fig:latent_cae}
\end{subfigure}
\hfill
\begin{subfigure}{0.49\linewidth}
    \centering
    \includegraphics[width=\linewidth, keepaspectratio]{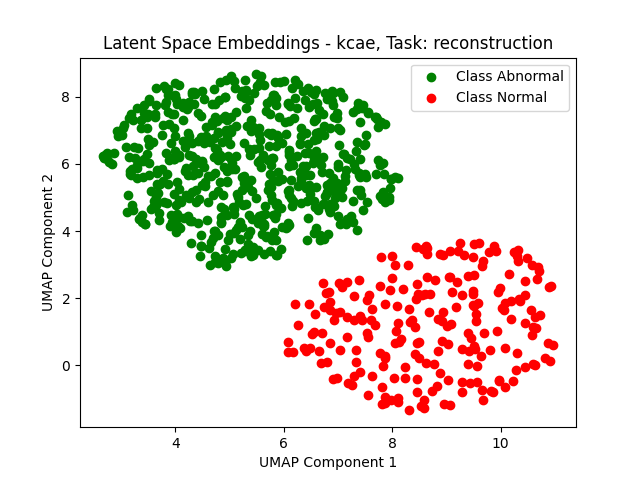}
    \caption{}
    \label{fig:latent_kcae}
\end{subfigure}
\centering
\caption{Latent space for the reconstruction task learned by  CAE (a) and KCAE (b) models.}
\end{figure}

The latent space distribution obtained from the KCAE model is shown in Figure~\ref{fig:latent_kcae}. The UMAP projection reveals a clear separation between the abnormal and normal heartbeat instances, with both classes forming distinct, though relatively diffuse, clusters. This indicates that the KCAE is able to extract meaningful, class-relevant features even in the absence of supervision (since the reconstruction task is unsupervised). The dispersion of points within each class cluster suggests a greater flexibility in the learned representations, which could potentially enhance robustness to input variability but may also allow more overlap in certain regions.

Conversely, the CAE model produces more compact and tightly clustered latent embeddings, as depicted in Figures ~\ref{fig:latent_cae} and \ref{fig:latent_kcae}.

The normal and abnormal classes are not only well-separated but also demonstrate higher intra-class compactness compared to KCAE. This suggests that CAE may learn more uniform reconstructions with less internal variance. However, the more rigid encoding could be less adaptive in scenarios with high intra-class variability. Overall, both models demonstrate the ability to separate classes in the latent space, with CAE favoring compact, tightly clustered embeddings and KCAE promoting more distributed but separable representations.

\subsection{Reconstruction Loss Analysis}

\begin{figure}[ht]
\begin{subfigure}{\linewidth}
    \centering
    \includegraphics[width=\linewidth, keepaspectratio]{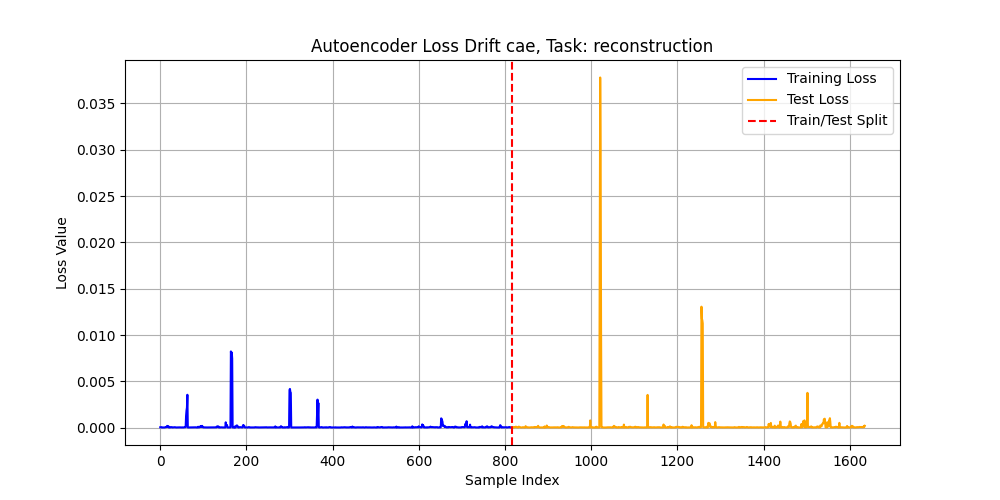}
    \caption{}
    \label{fig:drift_cae}
\end{subfigure}
\hfill

\begin{subfigure}{\linewidth}
    \centering
    \includegraphics[width=\linewidth, keepaspectratio]{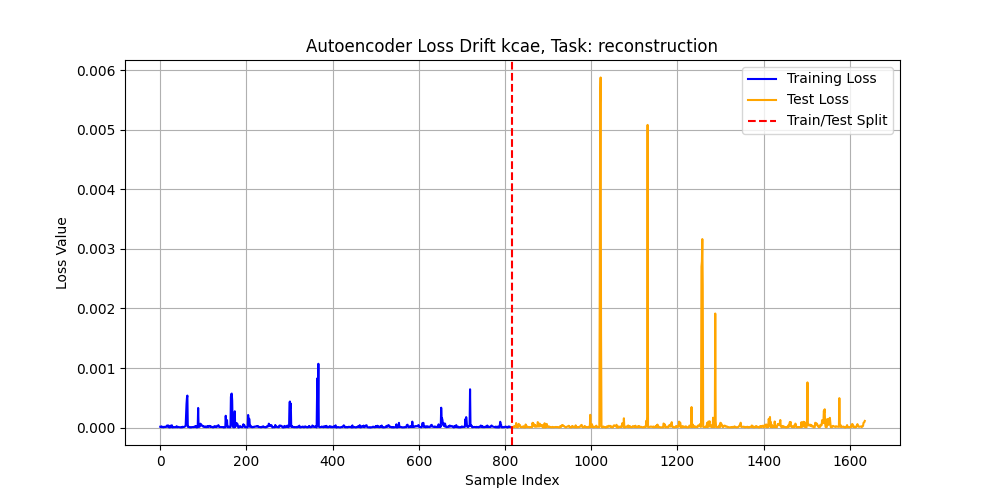}
    \caption{}
    \label{fig:drift_kcae}
\end{subfigure}
\centering
\caption{Loss drift analysis for CAE (A) and KCAE (B) models.}
\end{figure}

We examined the reconstruction behavior of CAE and KCAE models by analyzing loss distributions across training and test samples. The loss drift plots (Figure~\ref{fig:drift_kcae} for KCAE and Figure~\ref{fig:drift_kcae} for CAE) reveal how effectively each model generalizes to unseen data and highlight potential overfitting patterns.

\paragraph{KCAE Performance}

The KCAE model achieves consistently low reconstruction loss on training samples, with values near zero and only occasional minor spikes. On test data, reconstruction loss increases for specific samples but remains generally low (typically under 0.006). Most test samples maintain small reconstruction errors, indicating strong generalization capability, though the model shows sensitivity to outliers and difficult examples.

\paragraph{CAE Performance}

The CAE model shows comparable training performance with low reconstruction error and intermittent spikes. However, test set performance reveals greater variability, with multiple pronounced peaks exceeding 0.035 . These frequent, large spikes suggest the CAE has more difficulty with certain unseen instances and greater susceptibility to loss drift when encountering samples outside the training distribution.

\paragraph{Comparative Analysis}

Both models achieve low training loss, but KCAE demonstrates superior test performance with lower and more consistent reconstruction errors. This enhanced generalization likely stems from kernel-based convolutions enabling more adaptable feature representations. The reduced variance in KCAE's loss distribution indicates greater robustness across diverse test samples, making it more practical for real-world applications involving novel or noisy data.

\begin{figure}[ht]
\begin{subfigure}{\linewidth}
    \centering
    \includegraphics[width=\linewidth, keepaspectratio]{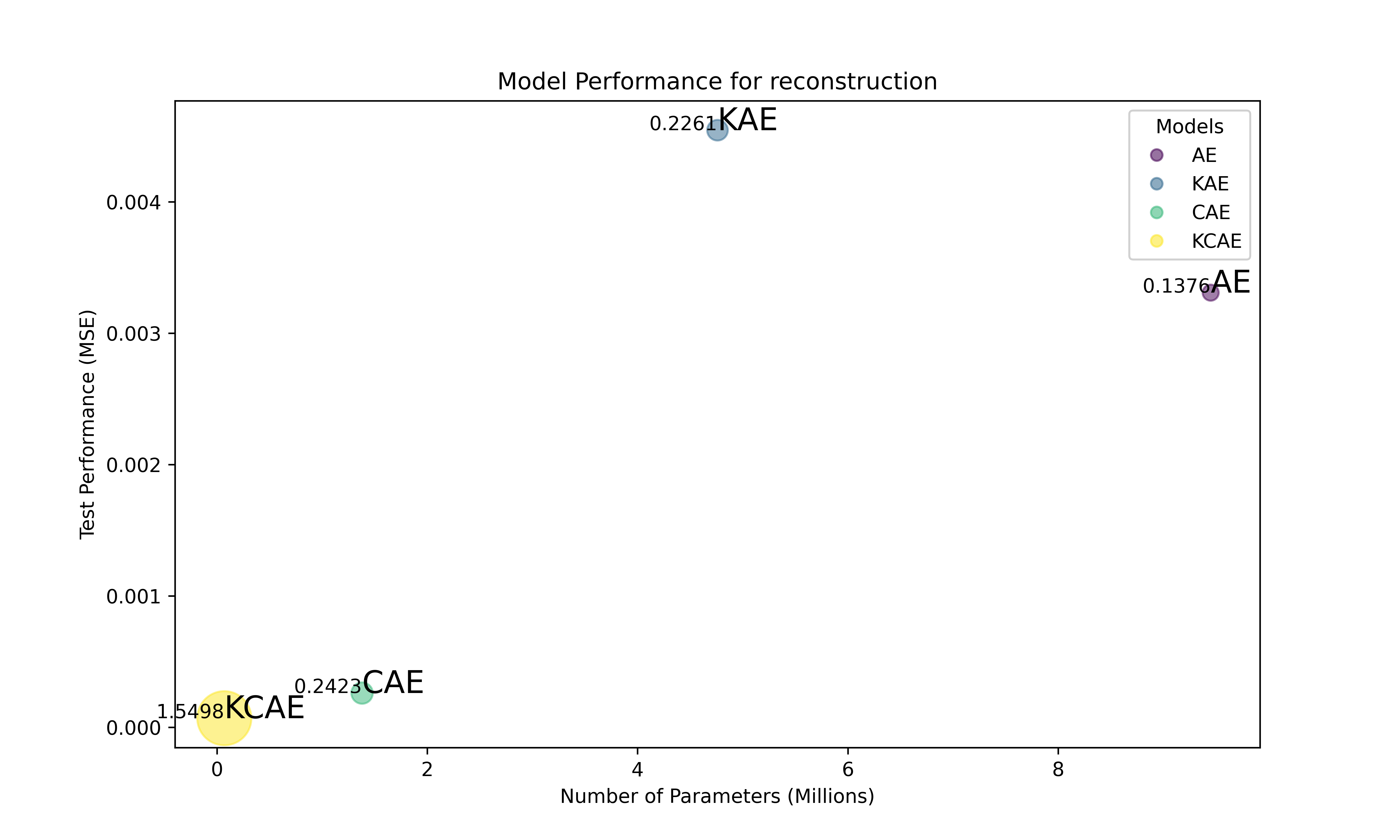}
    \caption{}
    \label{fig:performance_rec}
\end{subfigure}
\hfill
\centering
\caption{Efficiency (A)  evaluation for AE, KAE, CAE and KCAE autoencoders.}
\end{figure}

\paragraph{Model Performance}
Figure~\ref{fig:performance_rec} presents the trade-off between model complexity and reconstruction performance for the four evaluated autoencoder architectures: AE, KAE, CAE, and KCAE. The x-axis denotes the number of trainable parameters (millions), serving as a complexity metric, while the y-axis shows test reconstruction error measured by Mean Squared Error (MSE). Bubble sizes encode additional performance factors, potentially reflecting latent space quality or reconstruction robustness.
The fully connected architectures (AE and KAE) occupy the high-complexity region, with AE containing over 8 million parameters and KAE approximately 4 million. Despite substantial parameter counts, these models exhibit relatively poor reconstruction performance, achieving MSE values of 0.1376 (AE) and 0.2261 (KAE). This indicates that for time series reconstruction tasks, increased model capacity through dense layers does not guarantee improved generalization and may promote overfitting or inefficient representation learning.
Conversely, the convolutional architectures—CAE and KCAE—achieve superior reconstruction performance with significantly reduced parameter requirements. CAE attains an MSE of 0.2423 using approximately 1.5 million parameters, while KCAE achieves the optimal reconstruction error of 0.15498 with an even more compact design. KCAE's position in the bottom-left region of the plot demonstrates the ideal combination of low reconstruction error and minimal model complexity.
These results demonstrate that kernel-based convolutional architectures effectively capture essential time series features while maintaining computational efficiency. The performance differential between KCAE and KAE emphasizes the synergistic benefits of combining local receptive fields (convolutions) with adaptive kernel-based nonlinearities, enabling superior generalization with reduced complexity. The findings underscore the practical advantages of convolutional designs—particularly kernel-enhanced variants—for time series reconstruction applications where computational efficiency is relevant.

\section{Conclusion}
\label{sec:conclusion}

In this work, we conducted a comprehensive comparison of four autoencoder architectures—AE, KAE, CAE, and KCAE—on the AbnormalHeartbeat time series dataset, focusing on their ability to reconstruct heartbeat signals and generalize to unseen data. Through latent space analysis, loss drift evaluation, and quantitative performance metrics, we assessed both the representational capacity and reconstruction quality of these models.

Our results demonstrate that convolutional architectures (CAE and KCAE) consistently outperform their fully connected counterparts (AE and KAE) in terms of both reconstruction accuracy and latent space separability. Specifically, both CAE and KCAE produced well-separated latent embeddings for normal and abnormal heartbeats, with KCAE offering a more flexible but still clearly discriminative representation. The fully connected models exhibited higher intra-class variance and less compact latent spaces, particularly in the case of KAE, which suggests potential inefficiencies in learning meaningful features for time series data.

Loss drift analysis further revealed that KCAE achieved lower and more stable reconstruction loss on unseen test samples, with fewer high-error outliers compared to CAE. While both convolutional models generalized better than the fully connected ones, KCAE demonstrated superior robustness and generalization capacity, likely due to the inclusion of kernel-based nonlinearities enhancing local feature extraction.

Finally, when comparing test MSE against model complexity, KCAE stood out by achieving the best trade-off: the lowest reconstruction error with the smallest number of parameters. This finding highlights the efficiency of kernelized convolutional architectures in modeling complex temporal patterns in biomedical signals. Overall, our results suggest that KCAE offers a compelling balance of accuracy, robustness, and computational efficiency, making it a promising candidate for real-world applications involving unsupervised time series learning.

\section*{Acknowledgments}
This was was supported in part by FAIR Project.


\bibliographystyle{unsrt}  
\bibliography{biblio}  

\end{document}